\documentclass[runningheads]{llncs}

\usepackage{eccv}

\usepackage{eccvabbrv}
\usepackage{graphicx}
\usepackage{booktabs}

\usepackage[accsupp]{axessibility}  

\usepackage{multirow}
\usepackage{graphicx}
\usepackage{wrapfig}
\usepackage{colortbl}
\usepackage{algorithm}
\usepackage{algorithmic}
\usepackage{comment}
\usepackage{xcolor}
\usepackage{amsmath}
\usepackage{hyperref}       
\definecolor{link}{RGB}{225,0,0}
\definecolor{cite}{RGB}{218,112,214}
\hypersetup{
    colorlinks=true,
    linkcolor=link,
    citecolor=cite}

\usepackage{pifont}

\definecolor{my_green}{RGB}{51,102,0}
\definecolor{my_yellow}{RGB}{255,165,0}
\definecolor{my_red}{RGB}{204, 0, 0}
\newcommand{\colorcmark}{\textcolor{my_green}{\ding{52}}}
\newcommand{\colorxmark}{\textcolor{my_red}{\ding{55}}}
\usepackage{booktabs}    
\usepackage{multirow}    
\usepackage{subcaption}  
\usepackage{marvosym}

\usepackage{hyperref}

\usepackage{orcidlink}

\begin{document}

\title{SpatialEdit: Benchmarking Fine-Grained Image Spatial Editing} 

\titlerunning{SpatialEdit}

\author{Yicheng Xiao\inst{1,2,3}\thanks{Equal contribution. \Letter Corresponding author.}\and Wenhu Zhang\inst{4,2} \textsuperscript{\ensuremath{\star}}\and Lin Song\inst{2}\Letter\and Yukang Chen\inst{5}\and Wenbo Li\inst{2}\and Nan Jiang\inst{2}\and Tianhe Ren\inst{1}\and Haokun Lin\inst{2}\and Wei Huang\inst{1}\and Haoyang Huang\inst{2}\and Xiu Li\inst{3}\and Nan Duan\inst{2}\and Xiaojuan Qi\inst{1}\Letter}

\authorrunning{Y.~Xiao et al.}

\institute{The University of Hong Kong \and
JD Explore Academy \and Tsinghua University \and The Hong Kong University of Science and Technology \and The Chinese University of Hong Kong
}

\maketitle
\begin{figure}[ht]
  \centering
  \includegraphics[width=\linewidth]{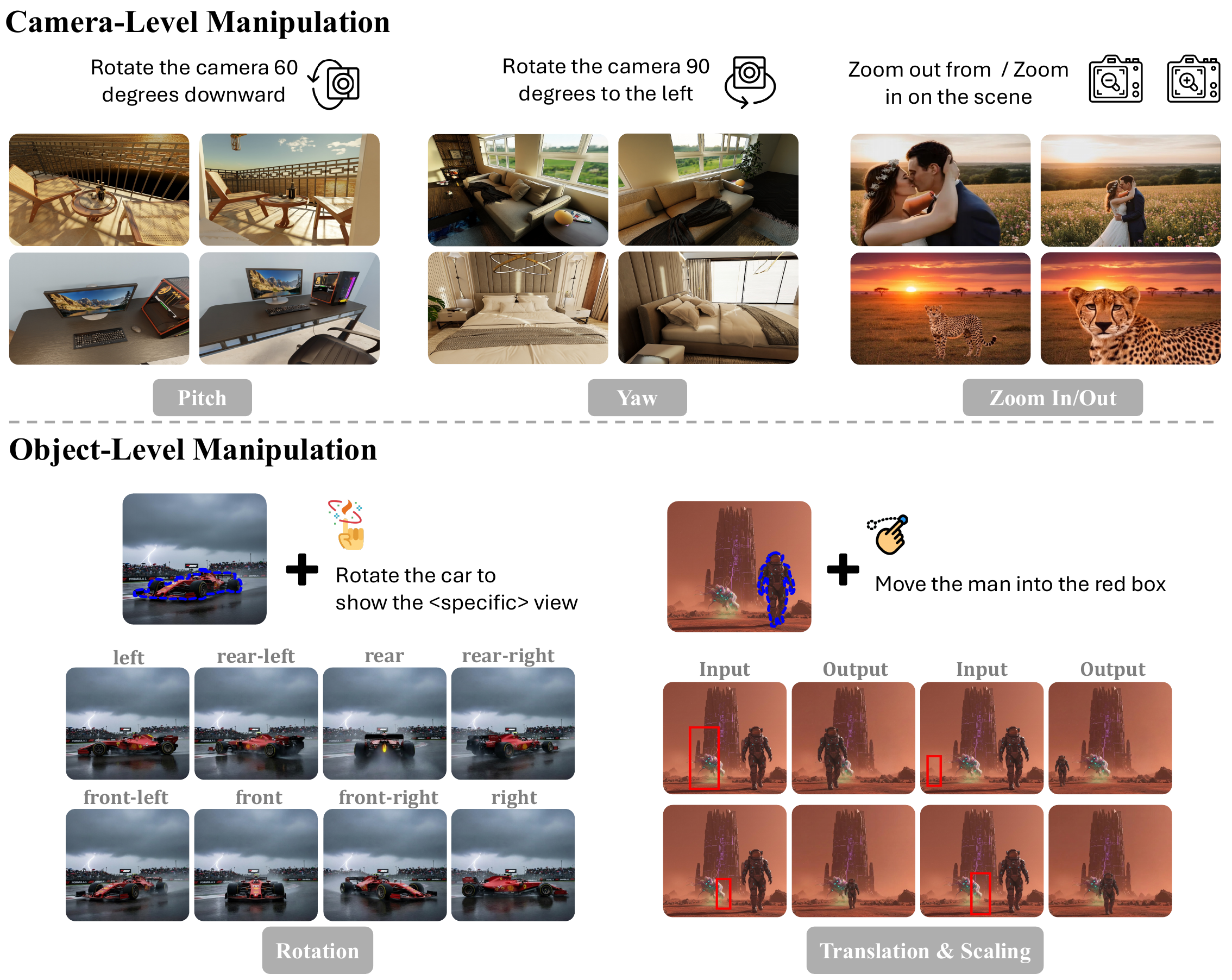}
  \caption{\textbf{Illustration for image spatial editing}. It comprises two components: (1) camera-centric view manipulation, including pitch, yaw, and zoom transformations; and (2) single-object manipulation, encompassing object rotation while preserving the background, as well as translation and scaling of objects specified via user-defined bounding boxes.
  }
  \label{fig:task_definition}
  \vspace{-20pt}
\end{figure}
\begin{abstract}
Image spatial editing performs geometry-driven transformations, allowing precise control over object layout and camera viewpoints. Current models are insufficient for fine-grained spatial manipulations, motivating a dedicated assessment suite.
Our contributions are listed: (i) We introduce SpatialEdit-Bench, a complete benchmark that evaluates spatial editing by jointly measuring perceptual plausibility and geometric fidelity via viewpoint reconstruction and framing analysis.
(ii) To address the data bottleneck for scalable training, we construct SpatialEdit-500k, a synthetic dataset generated with a controllable Blender pipeline that renders objects across diverse backgrounds and systematic camera trajectories, providing precise ground-truth transformations for both object- and camera-centric operations.
(iii) Building on this data, we develop SpatialEdit-16B, a baseline model for fine-grained spatial editing. Our method achieves competitive performance on general editing while substantially outperforming prior methods on spatial manipulation tasks.
All resources will be made public at \url{https://github.com/EasonXiao-888/SpatialEdit}.

\end{abstract}
\section{Introduction}
\label{sec:intro}
\begin{table*}[b]
    \centering
    \scriptsize
    \renewcommand\arraystretch{1.0}
    \setlength{\tabcolsep}{5pt}
    \vspace{-12pt}
    \begin{tabular}{l|ccccccc}
    \toprule
    \multirow{2}{*}{Benchmark} & Object & Object &  Object & Camera & VLM & Precise\\
    & Translation & Scaling  & Rotation & Manipulation & Metric & Metric\\
    \midrule
    ImgEdit~\cite{imgedit} & \colorxmark & \colorxmark & \colorxmark  & \colorxmark &\colorcmark & \colorxmark \\
    GEdit~\cite{Step1X-Edit} & \colorcmark & \colorxmark & \colorxmark  & \colorxmark & \colorcmark &\colorxmark\\
    CEdit~\cite{longcat} & \colorcmark & \colorxmark & \colorxmark  & \colorcmark & \colorcmark & \colorxmark \\
    \rowcolor{gray!10} SpatialEdit-Bench & \colorcmark & \colorcmark & \colorcmark  & \colorcmark & \colorcmark & \colorcmark \\
    \bottomrule
    
    \end{tabular}
    \caption{\textbf{Characteristics comparison with other editing benchmarks.}}
    \label{tab:bench_compare}
\end{table*}
Modern image editing ~\cite{nano-banana, gpt4o, seedream4}  is quickly moving beyond what to change (\textit{e.g.}, add, remove, replace and style) toward \textit{where} and \textit{how} to change it in 3D space. We refer to this capability as image spatial editing: editing an image by applying geometry-driven transformations rather than appearance edits.
Concretely, spatial editing spans two complementary axes (\cref{fig:task_definition}): camera-centric view manipulation (\textit{e.g.}, yaw, pitch, and zoom) and object-centric manipulation (\textit{e.g.}, translate/scale an object within a user-specified box, or rotate an object to a desired canonical view).
This functionality is increasingly central to world-modeling and embodied perception pipelines, where controllable viewpoint change and object reconfiguration are prerequisites for interactive content creation, simulation, and downstream 3D reasoning. 

Despite rapid progress in image generation following user instructions~\cite{nano-banana, GPT-IMAGE-1, longcat, qwenimage}, precise spatial control remains brittle.
Existing systems fall into three common failure modes. First, many spatially-conditioned world-model or video-generation pipelines require expert inputs such as full 6-DoF camera trajectories~\cite{ motionctrl,Camclonemaster}, creating a steep usability barrier for typical image-editing scenarios.
Second, strong general-purpose instruction-based editors often excel at semantic edits ~\cite{Step1X-Edit, qwenimage, dreamomni}, but frequently miss metric or viewpoint intent-- e.g., ``rotate the camera $90^{\circ}$ to the right'' or ``rotate the object to show its front-right side''-- yielding outputs that look plausible but are spatially incorrect.
Third, some methods~\cite{recammaster, longcat, uniworld-v2} incorporate spatial reasoning but are typically narrow (one operation or one setting) and do not generalize across the diverse operation set demanded by real users. Together, these limitations suggest a gap between ``semantic alignment'' and faithful geometric compliance.

A key reason this gap persists is that evaluation for spatial editing remains underdeveloped as shown in~\cref{tab:bench_compare}. When metrics cannot reliably distinguish ``looks right'' from ``is right,'' model iteration becomes noisy and progress is hard to measure.
To address this, we introduce SpatialEdit-Bench, a benchmark that covers both object-level and camera-level spatial editing, together with geometry-aware evaluation tailored to viewpoint changes.
Beyond detector-driven composition and framing analysis (our Framing Error: FE), we quantify Viewpoint Error (VE) by reconstructing the camera pose in 3D space, enabling a direct check of whether the edited result matches the intended geometric transformation.
In controlled validation with fine-grained pose variations, these metrics show substantially higher reliability than vision-language-based judging used in prior work~\cite{longcat, Step1X-Edit}, underscoring the necessity of geometry-sensitive evaluation for diagnosing true spatial capability.

\begin{figure}[tb]
  \centering
  \includegraphics[width=1.0\linewidth]{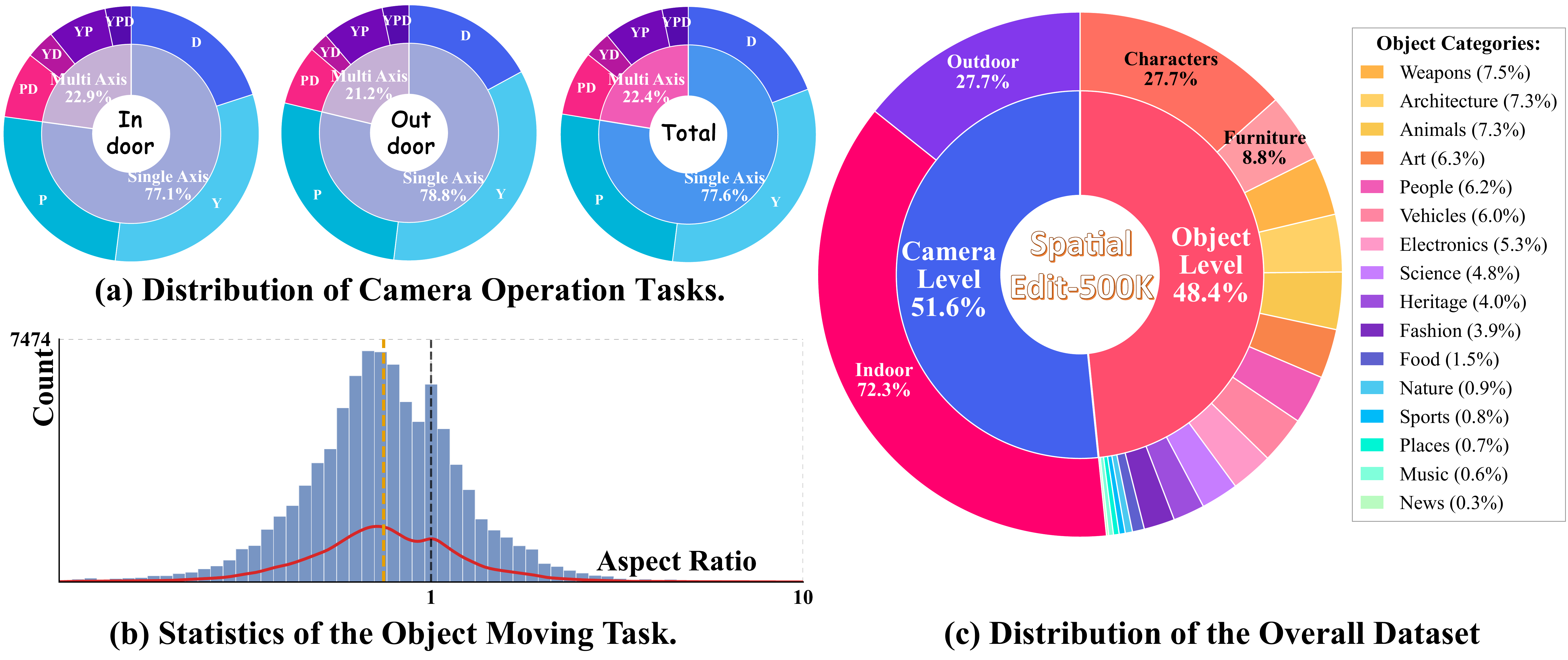}
  \caption{\textbf{Statistics of SpatialEdit-500k}. (a) Distribution of camera-level data across seven sub-tasks in outdoor and intdoor scenes, where Y, P, and D denote Yaw, Pitch, and Distance, respectively. (b) Aspect ratio distribution of bounding boxes for the moving task at the object level. (c) Object category statistics across the entire dataset.
  }
  \label{fig:task_distribution}
\end{figure}

Benchmarking alone, however, is not enough-- \textbf{training data is the real bottleneck for fine-grained spatial editing}. Such data is difficult to obtain at scale because it requires paired images with known geometric transformations, consistent object identity across edits, and faithful, unambiguous instructions, all while covering a wide range of scenes, object categories, and camera configurations.
To address this, we build a scalable and controllable data engine in Blender~\cite{blender} to synthesize paired supervision together with corresponding textual instructions. For object-level spatial editing, we render a large collection of GLB assets from eight preset viewpoints to generate source images. We then use VLMs~\cite{gpt4.1, gemini2.5} to verify the availability of a front view and assign object names, while SAM3~\cite{sam3} segments each object to produce mask labels. Next, we generate diverse backgrounds with a high-quality text-to-image model~\cite{nano-banana} and inpaint the rendered object into these backgrounds, producing realistic edited images with ground-truth spatial intent.
For camera-level editing, we curate a rich set of indoor and outdoor scenes, select salient objects as focal targets, and systematically sample camera poses around them by varying yaw, pitch, and zoom. We further use a VLM to generate paired images with accurate viewpoint changes and to produce natural language instructions that support flexible camera-centric edits.
As shown in \cref{fig:task_distribution}, the resulting SpatialEdit-500k achieves high diversity and a well-balanced distribution across task types. 
Building on this data, we develop SpatialEdit-16B, a fine-grained spatial editing model that combines a pretrained multimodal encoder~\cite{qwen3vl} with an MM-DiT decoder ~\cite{stablediffusion3}. We first ensure strong general editing behavior via pretraining on open-source editing data~\cite{gpt-image-edit}, and then specialize using parameter-efficient fine-tuning (LoRA) on SpatialEdit-500k. Empirically, this strategy yields a model that is competitive on general editing benchmarks while substantially improving on spatial tasks across object manipulation and camera control-precisely the regime where prior systems most often fail. 

To evaluate our approach, we use both the publicly available GEdit~\cite{Step1X-Edit} and our newly introduced SpatialEdit-Bench for general and spatial editing tasks.
Specifically, while maintaining comparable performance on general editing (7.52 on GEdit-Bench), our method acquires precise editing capabilities through continued training (as shown in~\cref{fig:task_definition}), surpassing the current open-source state-of-the-art model, LongCatImage-Edit, by 0.300 and 0.127 points on moving and rotation scores, respectively, while achieving the lowest error in camera control.
Furthermore, video-based world models~\cite{kling, vidu} remain significantly inferior to image-based spatial editing models in performing fine-grained spatial manipulation guided by text instructions.
Additionally, our model can also serve as a practical enhancement tool for single-view reconstruction.
\vspace{-10pt}
\section{Related Work}

\vspace{0.05in}\noindent\textbf{Image Editing and Generative Models.}
Diffusion-based generative models~\cite{cao2025hunyuanimage,qwenimage,seedream2025seedream} have greatly improved the fidelity and controllability of image editing. Instruction-based editing~\cite{bagel,emu-edit,llmga} modifies images according to natural language instructions while preserving overall semantics, but relies on large-scale instruction-faithful supervision. Early pipelines such as InstructP2P~\cite{instructp2p} combine prompt engineering with diffusion editing operators like Prompt-to-Prompt~\cite{p2p} built on latent diffusion~\cite{LDM}, while later works expand training data through human edits, inpainting-based synthesis, compositing, and expert-model orchestration~\cite{magicbrush,dreamomni,omniedit,seed-edit}. Recent unified frameworks further support richer instructions and multi-task editing~\cite{dreamve,dreamomni}.
Beyond text-only conditioning, reference-based methods improve controllability~\cite{dreambooth,textual-inversion}, whereas later methods introduce lightweight visual adapters~\cite{ip-adapter,blipdiffusion}.
More recent systems encode reference images as visual tokens within DiT~\cite{DIT,iclora,ominicontrol,kontext,qwenimage,omnigen,dreamomni}, with transformer-based backbones such as DiT improving scalability and conditioning flexibility~\cite{DIT}.

\vspace{0.05in}\noindent\textbf{Spatially-Aware Visual Manipulation.}
Recent progress in spatially conditioned generative modeling shows that explicit spatial control can be learned at scale.
Prior work has explored controllable viewpoint manipulation through camera-motion or 6-DoF conditioning (MotionCtrl~\cite{motionctrl}, CameraCtrl~\cite{cameractrl}, CVD~\cite{CVD}), camera-aware DiT architectures (AC3D~\cite{ac3d}), and more general control interfaces (OminiControl~\cite{ominictrl}).
Others incorporate geometric cues such as dense point tracks~\cite{cotracker,spatialtracker}, enabling geometry-aware generation in GS-DiT~\cite{gs-dit} and Diffusion-as-Shader~\cite{das}. Camera trajectory manipulation and novel-view synthesis using simulated data (Kubric~\cite{kubric}) are explored in GCD~\cite{gcd}, while Recapture~\cite{recapture} studies adaptation-based camera control for real images. However, evaluation of spatial manipulation remains limited, as existing benchmarks rely on coarse metrics or semantic alignment checks. We address this by introducing a unified benchmark with ground-truth geometric annotations and geometry-aware metrics that explicitly measure transformation accuracy and viewpoint correctness.
\section{Image Spatial Editing}

\begin{figure}[tb]
  \centering
  \includegraphics[width=\linewidth]{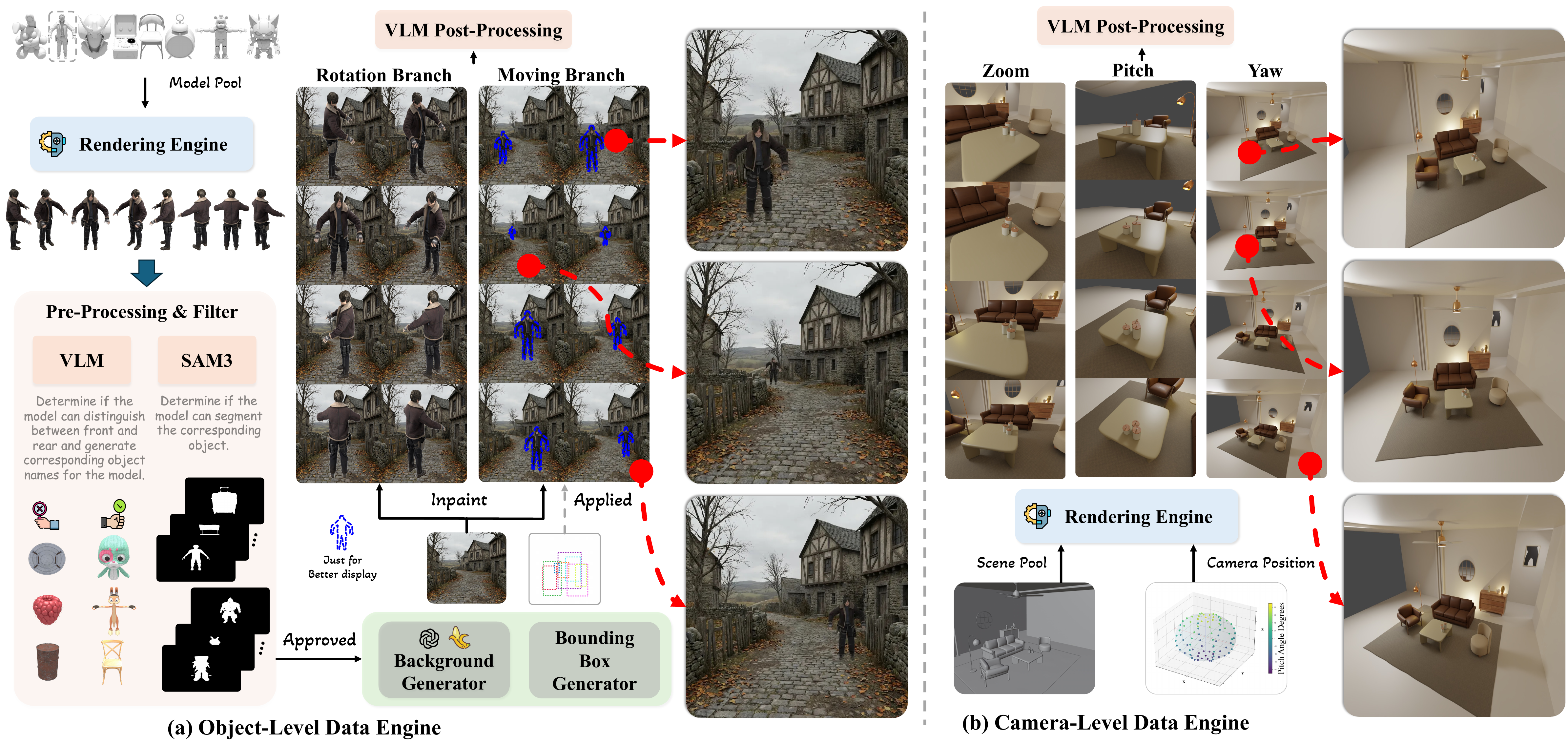}
  \caption{\textbf{SpatialEdit-500k data generation pipeline}. We leverage Blender to synthesize both objects and scenes, while preprocessing 3D assets using SAM3 and a vision-language model. The object-level engine constructs two inpainting-based data branches to generate object transformations, including rotation, translation, and scaling. The camera-level engine produces viewpoint transformation data by sampling different camera poses, resulting in variations in yaw, pitch, and zoom.
  }
  \label{fig:data_engine}
\end{figure}

\subsection{Revisiting Image Spatial Manipulation}
Image spatial manipulation has traditionally been formulated with explicit geometric constraints (e.g., view synthesis and pose-conditioned generation), but modern instruction-following editing models are expected to perform it from language supervision alone, exposing a key mismatch between semantic alignment and geometric compliance.
In practice, outputs often look plausible yet violate metric intent, especially for fine-grained camera operations (yaw/pitch/zoom) and canonical object reorientation.
Camera-centric view manipulation requires globally coherent re-projection and framing consistency, whereas object-centric manipulation demands localized, identity-preserving transformations disentangled from the background.
This motivates revisiting spatial manipulation as a first-class image editing capability with (i) a unified task definition spanning camera- and object-level control, (ii) geometry-aware evaluation that can distinguish “looks right” from “is right” (e.g., viewpoint- and framing-sensitive metrics), and (iii) scalable paired supervision with unambiguous transformation intent.
Consequently, it directly guides our benchmark, data engine, and model design.

\subsection{Task Definition}
To bridge the gap between semantic intent and geometric precision, we categorize spatial editing into two primary axes: object-centric manipulation and camera-centric view control.
\paragraph{\textbf{Object-Level Spatial Manipulation.}} We aim to edit individual objects within the image, including translation (repositioning), scaling (resizing), and rotation (orientation change) of specific entities while maintaining scene coherence. 
To enable granular control, we employ red bounding boxes to define the target translation and scaling operations. Users can constrain object movement and resizing either through textual instructions or by directly drawing a target rectangle on the canvas. For orientation, we discretize object orientation into eight canonical viewpoints: right, front-right, front, front-left, left, rear-left, rear, and rear-right.

\paragraph{\textbf{Camera-Level View Control.}}
This task involves manipulating the global imaging perspective to synthesize novel viewpoints without altering the underlying scene content. We parameterize this space through three degrees of freedom:(i) Pitch \& Yaw: We discretize vertical tilt (pitch) at 15° intervals and horizontal panning (yaw) at 45° increments. This grid provides comprehensive coverage of practical camera trajectories. (ii) Zoom: Focal length variations are modeled to simulate movement toward or away from the focal point.
By unifying these parameters, our framework elevates the editing task from a 2D image-to-image mapping to a geometry-aware transformation, effectively modeling the scene as a 3D environment with explicitly defined camera and object states.

\subsection{SpatialEdit-500k Dataset}
In this section, we will mainly introduce how we collect and create such a dataset to support image spatial editing.

\subsubsection{Object-Centric Data Generation Pipeline.}
To construct a high-quality object-centric dataset with controllable viewpoints and spatial variations, we design a multi-stage data curation pipeline. As shown in the left of ~\cref{fig:data_engine}, our pipeline progressively filters, augments, and annotates 3D assets to ensure geometric consistency, recognizability, and spatial diversity. The overall procedure consists of multiple stages, as described in the following.

We begin with variant GLB assets curated by TexVerse~\cite{textverse} and render them in Blender~\cite{blender} under a predefined canonical front-facing camera configuration, fixing camera intrinsics and object alignment to ensure consistent nominal frontal views.
To guarantee view correctness and remove ambiguous assets, we employ an advanced Vision-Language Model~\cite{gemini2.5} to verify that each rendered image corresponds to a valid frontal view and exhibits minimal side-view characteristics, discarding assets that fail these criteria.
For each retained GLB model, we render eight uniformly distributed viewpoints around the object while maintaining consistent camera intrinsics. We then apply the Segment Anything Model (SAM3) to obtain object masks for each view, using the first sentence of the \textit{TexVerse} caption as the textual prompt to verify correct object localization and segmentation; views failing this check are removed. To introduce spatial diversity, we generate eight additional renderings per valid view with randomized translations and scaling factors in Blender, and re-apply SAM3 to verify that the perturbed objects remain visible and properly contained within the image frame, retaining samples with at least one valid rendering.
For each canonical front view, we employ Nano-Pro~\cite{nano-banana} to synthesize a semantically compatible background image conditioned on the object's appearance. We composite these backgrounds with the validated multi-view images and their spatially perturbed variants, blending foreground objects while preserving geometric consistency. Finally, we project the known ground-truth 3D bounding boxes into the image plane to obtain precise 2D bounding box annotations for each transformed object.

\subsubsection{Camera-Centric Data Generation Pipeline.}
As depicted in the right pane of Fig.~\ref{fig:data_engine}, we established a high-fidelity 3D simulation environment to systematically sample camera trajectories for viewpoint manipulation.
We first build a large-scale pool of high-quality 3D scenes containing diverse indoor and outdoor layouts with semantically coherent object arrangements and physically plausible geometry. For each scene, we manually select $N_{\text{scene}}^{\text{target}}$ visually salient objects as camera focus targets, ensuring that the selected objects are recognizable and sufficiently exposed within the scene. These objects serve as anchors for defining camera viewpoint changes. We parameterize camera motion relative to the focus object using three degrees of freedom: yaw, pitch, and distance, corresponding to common camera operations such as horizontal orbiting, vertical tilting, and zooming. Using Blender~\cite{blender}, we systematically sample camera poses around each focus object by traversing predefined ranges of these parameters while keeping camera intrinsics fixed. Each sampled pose is used to render a candidate scene image, producing diverse viewpoints while preserving the underlying scene configuration and object layout.

To ensure dataset reliability, we apply a dual-branch quality filtering pipeline to remove invalid renderings. One branch employs a YOLO-based~\cite{yolov10} detector to verify the visibility of the focus object and discard images where the object is missing, severely occluded, or truncated. The other branch uses the vision-language model QwenVL-30B to assess semantic and geometric plausibility, filtering out renderings that exhibit mesh interpenetration, extreme or unnatural viewpoints, or visually meaningless scene compositions. For each valid rendering, we record the raw camera parameters $(\theta_y, \theta_p, d)$ and construct viewpoint pairs by sampling two poses associated with the same focus object, computing their relative transformation $(\Delta\theta_y, \Delta\theta_p, \Delta d)$. These transformations are first converted into templated camera-editing instructions describing the viewpoint change. We further provide human-like instructions alongside a concise target scene description, which serves as a prompt enhancer to reduce the difficulty of geometric editing. The resulting dataset contains high-quality image pairs with controlled camera transformations, associated camera parameters, and diverse natural language instructions, enabling systematic evaluation of camera-centric spatial editing capabilities.

\subsection{SpatialEdit-Benchmark}
To rigorously evaluate image spatial editing, we construct a benchmark focusing on spatial transformation tasks, jointly measuring geometric accuracy, semantic consistency, and structural preservation. Unlike conventional editing benchmarks centered on appearance changes, ours emphasizes spatial operations such as object scaling, translation, rotation, and camera viewpoint adjustment. 

\subsubsection{Evaluation Metric of Object-Level Task.}
Since we divide object-level manipulation into object moving and rotation, for the object moving branch, we can accurately evaluate it using only the detection model.
\paragraph{\textbf{Moving Score.}}
To quantitatively verify whether the predicted object location satisfies the prescribed absolute spatial constraint, we first employ a detection model to calculate the IoU.
However, geometric correctness alone is insufficient, as spatial translation may introduce semantic artifacts or contextual inconsistencies. Therefore, we further introduce a VLM to compute an object consistency score $S_{\mathrm{oc}} \in [0,1]$, which evaluates both subject integrity and environmental coherence after transformation.
Then the moving score is defined as:
\begin{equation}
\text{MS} = \sqrt{ \mathrm{IoU}({b}_\text{gt}, {b}_\text{pred}) \cdot S_{\mathrm{oc}} }.
\end{equation}
The geometric mean formulation enforces a multiplicative coupling between spatial accuracy and semantic fidelity, thereby penalizing imbalance.
\paragraph{\textbf{Rotation Score.}}
For the object rotation branch, $\widetilde{I_{\text{pred}}}$ denotes the image after applying a viewpoint transformation parameterized by yaw $\theta$ and pitch $\phi$.
Since rotation correctness is inherently viewpoint-sensitive and difficult to localize geometrically, we employ an advanced closed-source VLM~\cite{gemini2.5} to estimate the viewpoint correctness score $S_{\mathrm{view}}$, which measures whether the rendered perspective matches the specified angular configuration.
To further prevent appearance drift or structural distortion introduced during rotation, we additionally compute a consistency score $S_{\mathrm{cons}}$ by the same VLM.
The final rotation score is defined as:
\begin{equation}
\text{RS} = \sqrt{ S_{\mathrm{view}} \cdot S_{\mathrm{cons}}}.
\end{equation}
This multiplicative design also enforces simultaneous viewpoint correctness and semantic continuity, preventing trivial viewpoint hallucination that disregards object identity or scene plausibility.

\subsubsection{Evaluation Metric of Camera-Level Task.}
Given a triplet of source, ground-truth, and predicted views $(\mathcal{I}_{\text{src}}, \mathcal{I}_{\text{gt}}, \mathcal{I}_{\text{pred}})$, we aim to evaluate camera-level editing from two complementary aspects: (i) \textit{Framing Error (FE)} in the image plane (the focus object should remain visible with correct composition), and (ii) \textit{Viewpoint Error (VE)} in terms of camera extrinsics (the predicted camera pose should match the target pose up to scene scale).
We therefore adopt a dual-metric protocol, reporting a detector-based metric with YOLO~\cite{yolov10} and a geometry-aware metric with VGGT~\cite{wang2025vggt}.
\paragraph{\textbf{Viewpoint Error.}}
Specifically, to measure viewpoint correctness in a geometry-aware manner, we employ VGGT~\cite{wang2025vggt}, which is a feed-forward transformer that directly infers key 3D attributes of a scene, including camera parameters (intrinsics/extrinsics), depth maps, point maps, and 3D point tracks, with the input of a single or a set of images from various viewpoints. It models the scene in a globally consistent 3D representation rather than relying on purely 2D appearance cues.
In our evaluation, VGGT takes $(\mathcal{I}_{\text{src}}, \mathcal{I}_{\text{gt}}, \mathcal{I}_{\text{pred}})$ as inputs and returns estimated world-to-camera extrinsics:
\begin{equation}
(\widetilde{\mathcal{R}}_{\text{src}}, t_{\text{src}}),\;
(\widetilde{\mathcal{R}}_{\text{gt}}, t_{\text{gt}}),\;
(\widetilde{\mathcal{R}}_{\text{pred}}, t_{\text{pred}})
= f_{\text{VGGT}}(\mathcal{I}_{\text{src}}, \mathcal{I}_{\text{gt}}, \mathcal{I}_{\text{pred}})),
\end{equation}
from which we compute camera centers in world coordinates:
\begin{equation}
{\mathbf{C}} = -\widetilde{\mathbf{R}}^{\top} {\mathbf{t}}, \quad \widetilde{\mathbf{R}} = (\widetilde{\mathcal{R}}_{\text{src}}, \widetilde{\mathcal{R}}_{\text{gt}}, \widetilde{\mathcal{R}}_{\text{pred}}),\quad \mathbf{t}=(t_{\text{src}}, t_{\text{gt}}, t_{\text{pred}}).\end{equation}
We then calculate a baseline-normalized translation error (to remove sensitivity to global scene scale) and and a rotation error based on the geodesic distance on $\mathrm{SO}(3)$, which are formulated as:
\begin{equation}
\begin{aligned}
\epsilon_{\mathrm{xyz}}
&=
\frac{\left\lVert \mathcal{C}_{\mathrm{pred}} - \mathcal{C}_{\mathrm{gt}} \right\rVert_2}
{\left\lVert \mathcal{C}_{\mathrm{gt}} - \mathcal{C}_{\mathrm{src}} \right\rVert_2 + \varepsilon}, 
\\[6pt]
\epsilon_{\mathrm{rot}}
&=
\frac{1}{90}\,
d_{\mathrm{geo}}\!\left(
\widetilde{\mathcal{R}}_{\mathrm{pred}},
\widetilde{\mathcal{R}}_{\mathrm{gt}}
\right), 
\\[4pt]
d_{\mathrm{geo}}({x}_1, {x}_2)
&=
\arccos\!\left(
\frac{\mathrm{Tr}({x}_1^{\top}{x}_2) - 1}{2}
\right)
\cdot \frac{180}{\pi}.
\end{aligned}
\end{equation}
Finally, we aggregate them into a single pose error:
\begin{equation}
\text{VE}
=\frac{1}{2}\left(\epsilon_{\mathrm{xyz}}+\epsilon_{\mathrm{rot}}\right),
\end{equation}
where lower $\text{VE}$ indicates more accurate camera viewpoint editing.

\paragraph{\textbf{Framing Error.}}
Calculating with Viewpoint Error alone may not reflect whether the edited output preserves meaningful spatial layout under camera motion (e.g., objects might drift to incorrect positions or the scene structure may become distorted).
We thus introduce an object-centric spatial consistency metric (FE: Framing Error) based on detection.
We first introduce angle consistency.
Let $\mathcal{O}_{\text{gt}} = \{o_1^{\text{gt}}, \dots, o_n^{\text{gt}}\}$ and $\mathcal{O}_{\text{pred}} = \{o_1^{\text{pred}}, \dots, o_m^{\text{pred}}\}$ denote the sets of detected objects in $\mathcal{I}_{\text{gt}}$ and $\mathcal{I}_{\text{pred}}$, respectively.
For each object, we compute a ray direction~\cite{hartley,pinhole} from the image center to the object's bounding box center $(u, v)$:
\begin{equation}
\mathbf{r}(u,v) = \frac{[(u-c_x)/f,\ (v-c_y)/f,\ 1]^\top}{\|[(u-c_x)/f,\ (v-c_y)/f,\ 1]\|},
\end{equation}
where $(c_x, c_y)$ is the image center and $f$ is the focal length.
We establish correspondences between $\mathcal{O}_{\text{gt}}$ and $\mathcal{O}_{\text{pred}}$ via the Hungarian Matching~\cite{hungarian} algorithm, minimizing the sum of ray angles and area ratios.
Let $\mathcal{M} = \{(i,j)\}$ be the set of matched object pairs.
We compute the average ray angle difference $\epsilon_{\text{rag}}$ as:
\begin{equation}
\epsilon_{\mathrm{rag}} = \frac{1}{|\mathcal{M}|} \sum_{(i,j) \in \mathcal{M}} \arccos\!\bigl(\mathbf{r}_i^{\text{gt}} \cdot \mathbf{r}_j^{\text{pred}}\bigr) \times \frac{180}{\pi},
\end{equation}
where a lower value indicates better spatial alignment between the predicted and target object layouts.
Additionally, we verify whether the predicted image exhibits the correct scale change relative to the source for zoom editing commands.
Let $\mathcal{M}_{\text{zoom}}$ be matched object pairs between $\mathcal{I}_{\text{src}}$ and $\mathcal{I}_{\text{pred}}$.
Given command-specified distance change $\Delta d$ (negative for zoom-in, positive for zoom-out), we compute the zoom direction error as follows:
\begin{equation}
\epsilon_{\text{zde}} = \mathbb{I}\left[\widetilde{F^{\text{med}}_{i,j}} \left(\frac{1}{2}\log\!\left(\frac{|b_j^{\text{pred}}|}{|b_i^{\text{src}}|}\right)\right) \times \Delta d > 0\right],
\end{equation}
where $|b|$ denotes bounding box area, $\mathbb{I}[\cdot]$ is the indicator function and $\widetilde{F^{\text{med}}_{i,j}}$ indicates the median function.
This binary metric ensures that zoom-in commands ($\Delta d < 0$) produce larger objects ($s_{\text{log}} > 0$) and vice versa.
Finally, the framing error can be formulated as:
\begin{equation}
\text{FE}
=\frac{1}{2}\left(\epsilon_{\mathrm{rag}}+\epsilon_{\mathrm{zde}}\right).
\end{equation}

\section{Image Spatial Editing Model}
\begin{wrapfigure}{r}{0.45\textwidth}
    \centering
    \includegraphics[width=0.43\textwidth]{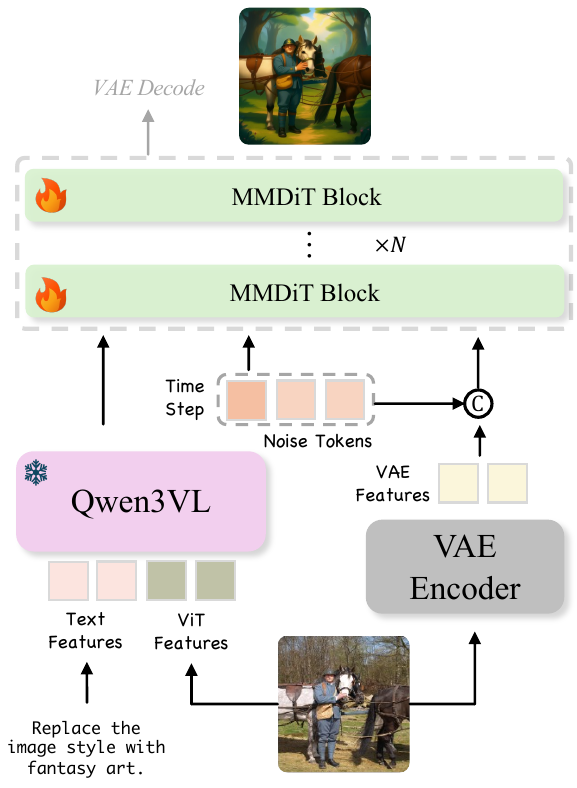}
    \caption{\textbf{Overview of SpatialEdit.}}
    \label{fig:wrap_example}
\vspace{-70pt}
\end{wrapfigure}

As shown in~\cref{fig:wrap_example}, we adopt a cascaded editing pipeline~\cite{qwenimage, mindomni, longcat}. Given an instruction and a reference image, a vision language model produces semantic embeddings as global conditioning. The image is encoded into a VAE latent, and an MMDiT~\cite{stablediffusion3} denoises it under multimodal guidance to obtain the edited latent, which is decoded by the VAE to the final output.Training proceeds in two stages: (1) adapt the model to image editing via fine-tuning on public editing data, and 
(2) specialize in image spatial editing scenario with LoRA post-tuning on our curated dataset, improving transformation control while preserving general priors.
\section{Experiments}
\begin{table}[tb]
\centering
\scriptsize
\renewcommand\arraystretch{1.0}
\setlength{\tabcolsep}{2.0pt}
\begin{tabular}{lcccccc}
\toprule
\multirow{3}{*}{Method} 
& \multicolumn{2}{c}{Object}  
& \multicolumn{2}{c}{Camera} & Object & Camera \\

\cmidrule(lr){2-3}
\cmidrule(lr){4-5}

& Moving 
& Rotation 
& Viewpoint 
& Framing 
& Overall 
& Overall   \\
& Score $\uparrow$
& Score $\uparrow$
& Error$\downarrow$
& Error$\downarrow$
& Score $\uparrow$
& Error $\downarrow$ \\

\midrule
\multicolumn{7}{c}{\textit{World Model}} \\
\midrule
ViduQ2-Turbo~\cite{vidu} 
& --    & --   
& 1.022 & 0.771 & --    & 0.897 \\
Kling-V2.5~\cite{kling}     
 & --    & --     
& 1.051 & 0.733 & -- &0.892\\
\midrule
\multicolumn{7}{c}{\textit{Closed-Source Image Model}} \\
\midrule
Nano-Banana~\cite{nano-banana}     
& 0.099 & 0.420 & 0.845 & 0.708 & 0.260 & 0.777 \\

Seedream4~\cite{seedream4}          
& 0.163 & 0.482 & 0.839 & 0.701 & 0.323 & {0.770} \\

\midrule
\multicolumn{7}{c}{\textit{Open-Source Image Model}} \\
\midrule
QwenImageEdit~\cite{qwenimage} 
 & 0.311 & 0.531 & 0.922 & 0.692 & 0.421 & 0.807 \\

Edit-R1~\cite{uniworld-v2}          
& 0.306 & 0.562 & 0.959 & 0.688 & 0.434 & 0.824 \\

LongCatImage-Edit~\cite{longcat}   
& 0.373 & 0.505 & 0.802 & 0.684 & {0.439} & 0.743 \\

\rowcolor{gray!10} 
SpatialEdit-PT (Baseline)      
& 0.186 & 0.489  & 0.890 & 0.719& 0.338 & 0.804 \\

\rowcolor{gray!10} 
\textbf{SpatialEdit}           
& \textbf{0.673} & \textbf{0.632} & \textbf{0.243} & \textbf{0.527} & \textbf{0.653} & \textbf{0.385} \\

\bottomrule
\end{tabular}
\vspace{4pt}
\caption{\textbf{Performance comparison on proposed SpatialEdit-Bench.}}
\label{tab:space_edit}
\vspace{-25pt}
\end{table}
\subsection{Training Details}
We pre-train the model on open-source editing datasets~\cite{gpt-image-edit} and proprietary internal data, explicitly excluding spatial editing samples (see supplementary for details). Training uses the AdamW~\cite{adamw} optimizer with $\beta_1=0.9$, $\beta_2=0.95$, a learning rate of $1\times10^{-4}$, and a linear warmup over the first 1,000 iterations.
\begin{figure}[t!]
  \centering
  \includegraphics[width=\linewidth]{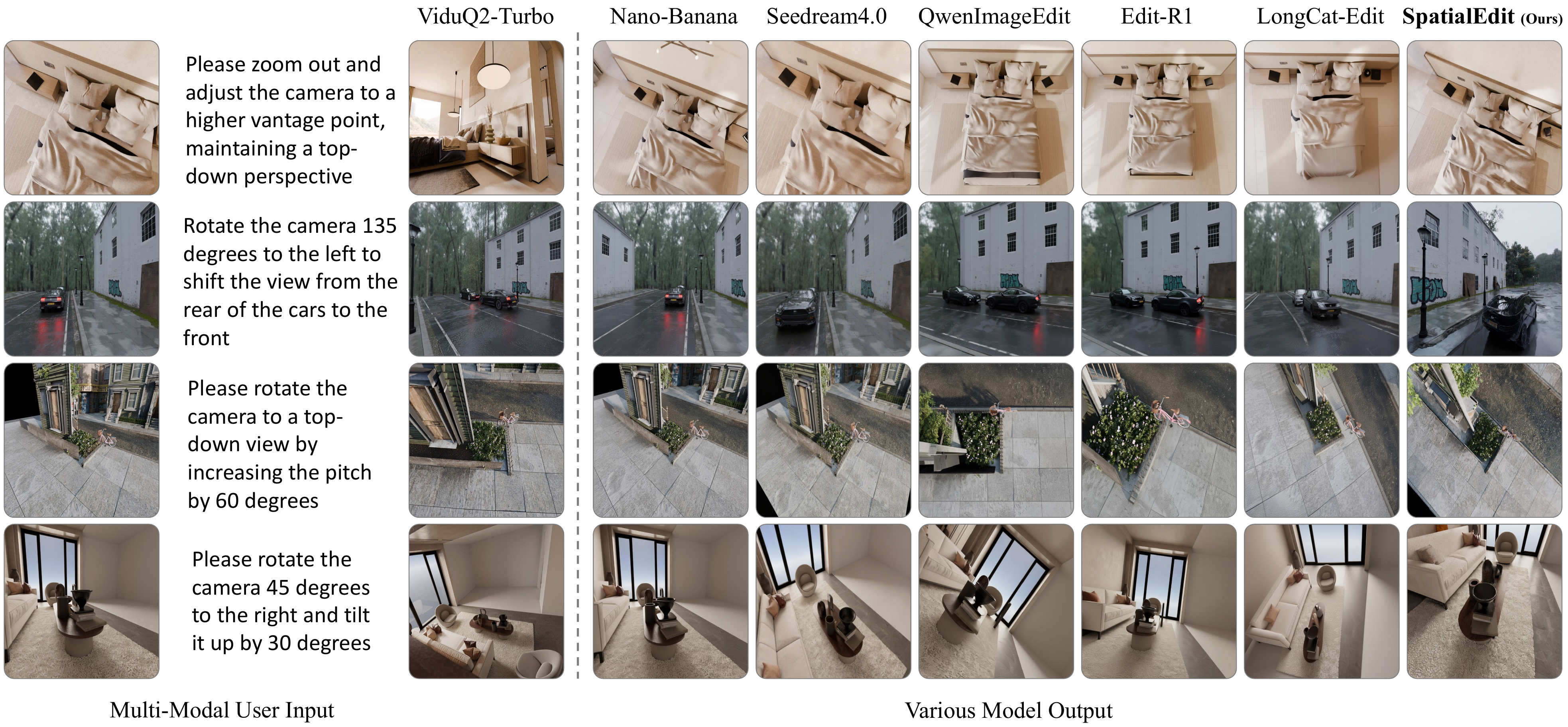}
  \caption{\textbf{Comparison of camera view manipulation across various methods}.
  }
  \label{fig:case_camera}
  \vspace{-10pt}
\end{figure}
\subsection{Quantitative Results}
\begin{figure}[th]
  \centering
  \includegraphics[width=\linewidth]{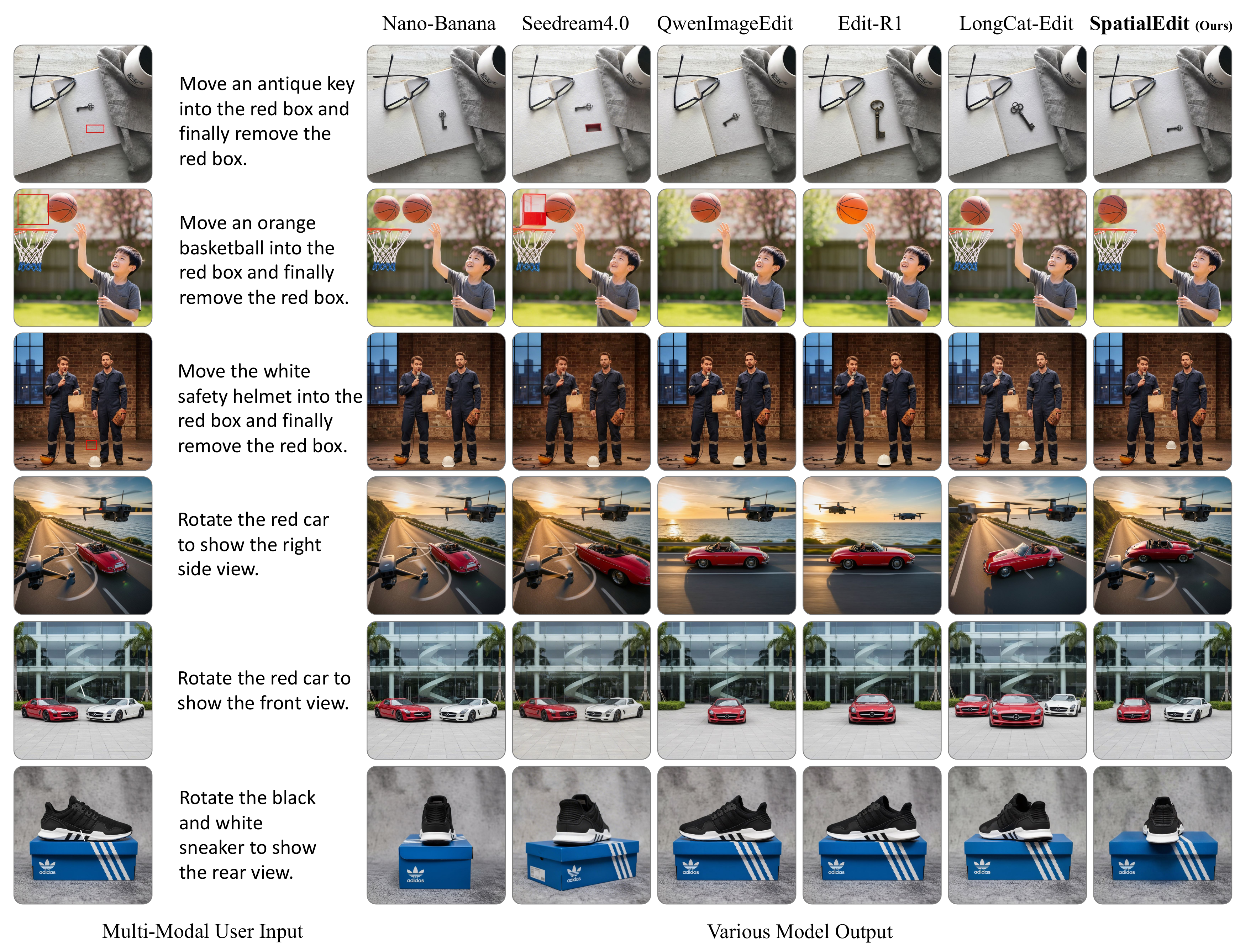}
  \caption{\textbf{Comparison of object-level manipulation across various methods}.
  }
  \label{fig:case_object}
\end{figure}
\paragraph{\textbf{Image Spatial Editing Performance.}}
As shown in~\cref{tab:space_edit}, our SpatialEdit achieves the best overall performance across both object-level and camera-level metrics.
For object-level tasks, our SpatialEdit significantly surpasses all baselines in object moving score with 0.673, while maintaining a competitive object rotation score (0.632).
On the other hand, SpatialEdit yields the lowest viewpoint error (0.243) and framing error (0.527), boosting the current SOTA method, LongCatImage-Edit by 0.358 in the overall camera error metric.
We further evaluate closed-source world models on precise camera viewpoint control from text instructions by sampling the final frame of the generated videos. The results show that their performance is weaker than that of mainstream image editing models, likely due to the challenge of maintaining consistent camera motion during video generation.

\paragraph{\textbf{General Editing Performance.}}
To validate the general editing capability of our model during pre-training, we evaluate it on GEdit, as shown in \cref{tab:gedit_bench}. Among open-source models, SpatialEdit achieves competitive performance, providing a strong foundation for subsequent fine-tuning on spatial editing tasks.

\subsection{Ablation Studies}
\begin{table*}[t]
\centering
\footnotesize
\begin{minipage}[]{0.44\textwidth}
\centering
\begin{minipage}[]{\textwidth}
\centering
\renewcommand\arraystretch{1.0}
\setlength{\tabcolsep}{0.8pt}
\begin{tabular}{cccccc}
\toprule
\multirow{2}{*}{Mov.} & \multirow{2}{*}{Rot.} & \multirow{2}{*}{Cam.} & Mov. & Rot. & Cam. \\
 &  &  & Score.$\uparrow$ & Score.$\uparrow$ & Error.$\downarrow$ \\
\midrule
\checkmark &  &  & 0.653  & -  & -\\
 & \checkmark & & - & 0.628 & - \\
 &  & \checkmark & - & - & 0.395 \\
\checkmark & \checkmark &  & 0.657 & 0.632 & -\\
\checkmark &  & \checkmark & 0.665 & - & 0.402 \\
\checkmark & \checkmark & \checkmark & 0.673 & 0.632 & 0.385 \\
\bottomrule
\end{tabular}
\caption{\textbf{The impact of training with different data combinations.} Mov, Rot, and Cam represent object moving, object rotation, and camera operation tasks, respectively.}
\label{tab:abl_stage}
\end{minipage}

\vspace{-3pt}  

\begin{minipage}[]{\textwidth}
\centering
\renewcommand\arraystretch{1.0}
\setlength{\tabcolsep}{3.0pt}
\begin{tabular}{cccc}
\toprule
& FE & VE & GPT4.1 \\
\midrule
Spearman Score & 0.659 & 0.932 & 0.445 \\
\bottomrule
\end{tabular}
\caption{\textbf{The effectiveness comparison across three type of metrics in camera evaluation using Spearman Correlation.}}
\label{tab:metric_ablate}
\end{minipage}

\end{minipage}%
\hspace{20pt}
\begin{minipage}[]{0.44\textwidth}
\centering
\renewcommand\arraystretch{1.0}
\setlength{\tabcolsep}{2.0pt}
\begin{tabular}{lccc}
\toprule
\multirow{2}{*}{Model} 
& \multicolumn{3}{c}{GEdit-Bench-EN$\uparrow$}  \\
\cmidrule(lr){2-4}
& SC & PQ & O \\
\midrule
\multicolumn{4}{c}{\textit{Closed Source}} \\
\midrule
Gemini 2.0~\cite{gemini2.5} & 6.73 & 6.61 & 6.32  \\
GPT Image 1~\cite{gpt-image-edit} & 7.85 & 7.62 & 7.53 \\
Nano Banana~\cite{nano-banana} & 7.86 & \textbf{8.33} & 7.54  \\
Seedream 4.0~\cite{seedream4} & \textbf{8.24} & 8.08 & \textbf{7.68} \\
\midrule
\multicolumn{4}{c}{\textit{Open Source}} \\
\midrule
UniWorld-v1~\cite{uniworld-v1} & 4.93 & 7.43 & 4.85 \\
MindOmni~\cite{mindomni} & 6.53 & 6.93 & 5.98 \\
OmniGen2~\cite{omnigen2} & 7.16 & 6.77 & 6.41 \\
FLUX.1 Kontext~\cite{flux.1_kontext} & 6.52 & 7.38 & 6.00  \\
BAGEL~\cite{bagel} & 7.36 & 6.83 & 6.52 \\
Step1X-Edit~\cite{Step1X-Edit} & 7.66 & 7.35 & 6.97 \\
Qwen-Image-Edit~\cite{qwenimage} & 8.00 & 7.86 & 7.56 \\
LongCat-Edit~\cite{longcat} & {8.18} & {8.00} & {7.64} \\
\rowcolor{gray!10} \textbf{SpatialEdit} & 8.09 & 7.80 & 7.52 \\
\bottomrule
\end{tabular}
\caption{\textbf{Performance comparison on GEdit-Bench~\cite{Step1X-Edit}}.}
\label{tab:gedit_bench}
\end{minipage}

\end{table*}
\paragraph{\textbf{Training data combinations.}}
As shown in~\cref{tab:abl_stage}, multi-task mixed training converges more reliably than single-task training.
Mov+Rot boosts both object metrics (0.657/0.632), and adding Cam further improves Mov and lowers camera error (Mov+Cam: 0.665, Cam: 0.402). Training on all three tasks yields the best overall trade-off (Mov: 0.673, Rot: 0.632, Cam: 0.385), indicating positive transfer from shared spatial supervision.
\paragraph{\textbf{Camera evaluation metrics.}}
We render $n$ fine-grained views of the same scene, fix one as the ground-truth, and treat the remaining views as pseudo edits with a known ordering. Each metric scores and ranks these views; we then compute Spearman correlation between the predicted and true rankings (Table~\ref{tab:metric_ablate}). VE attains the highest correlation, followed by FE, and both substantially outperform GPT, supporting the reliability of VE for camera evaluation.

\subsection{Qualitative Results}
\cref{fig:case_camera} compares camera view manipulations (zoom-out, yaw/pitch changes, and rotation/tilt).
SpatialEdit follows the requested viewpoint shifts more faithfully while better preserving scene geometry and reducing distortions (e.g., texture stretching, boundary drift, and hallucinations) than prior baselines.
Moreover, \cref{fig:case_object} evaluates object-level edits (moving items into a target region and rotating objects to specified views).
Competing methods often leave artifacts or alter the background, whereas SpatialEdit performs cleaner edits with higher object fidelity and stronger background preservation, yielding a better accuracy--preservation trade-off.

\subsection{Enhancement Tool for Single-view Reconstruction}
\begin{figure}[tb]
  \centering
  \includegraphics[width=\linewidth]{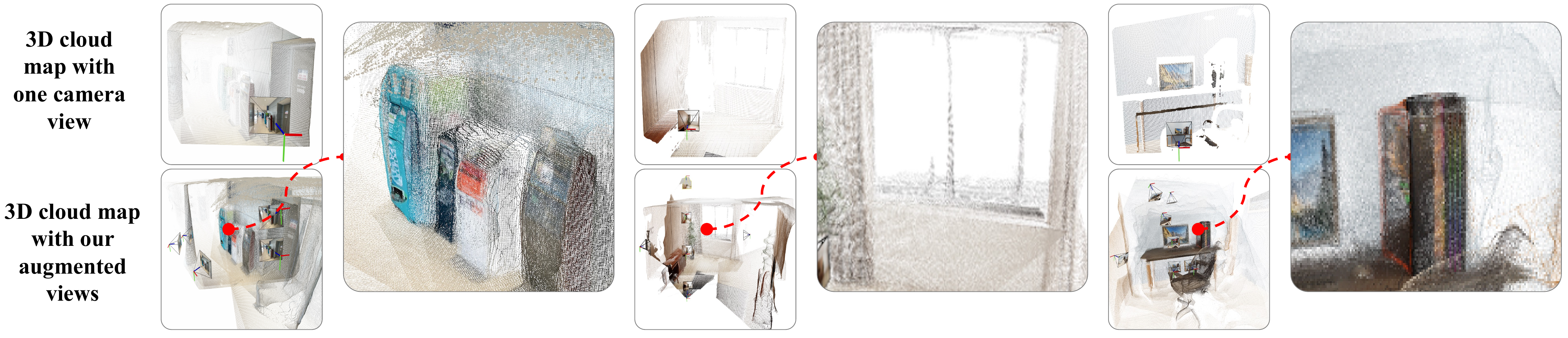}
  \caption{\textbf{Serving as an enhancement tool for single-view reconstruction}.
  }
  \label{fig:3d_recons}
\end{figure}
We propose a pipeline that uses SpatialEdit to improve 3D reconstruction when multi-view observations are unavailable. As shown in~\cref{fig:3d_recons}, single-view inputs suffer from depth-scale ambiguity and missing geometric cues. By editing the camera to synthesize novel viewpoints, our approach adds geometric constraints, leading to more accurate, structurally consistent, and detailed reconstructions.
\section{Conclusion}
We proposed a fine-grained spatial image editing paradigm, where edits are composed by object manipulations and explicit geometric control of camera viewpoints.
For rigorous assessment, we introduce SpatialEdit-Bench, which improves evaluation reliability by jointly measuring perceptual plausibility and geometric fidelity via viewpoint reconstruction and compositional analysis.
To mitigate the data bottleneck, we build SpatialEdit-500k, a controlled Blender-based dataset with diverse scenes and 3D assets.
Leveraging this data, we develop SpatialEdit-16B, a strong baseline that remains competitive on general editing while substantially advancing performance on challenging spatial manipulation tasks.
We hope our benchmark, dataset, and model will support reproducible progress and motivate future work that more tightly couples geometric inference with high-quality image synthesis.

\clearpage
\appendix
\section*{Supplemental Materials}
\section{Implementation Details}
We pre-trained the model using the open-source general image editing dataset gpt-image-edit and proprietary internal data, explicitly excluding all spatially edited samples. During pre-training, we used the AdamW optimizer with parameters $\beta_1=0.9$ and $\beta_2=0.95$, and a learning rate of $1\times10^{-4}$. A linear warm-up schedule was applied for the first 1000 iterations before transitioning to standard decay. In the post-training phase, we fine-tuned the model on our SpatialEdit-500k dataset using LoRA with rank 16 and $\alpha=16$, initializing the LoRA parameters with a Gaussian distribution.

\begin{table}[h]
\centering
\scriptsize
\renewcommand\arraystretch{1.0}
\setlength{\tabcolsep}{2.0pt}
\begin{tabular}{lccc}
\toprule
\multirow{3}{*}{Method} 
& \multicolumn{2}{c}{Camera}  & Camera \\

\cmidrule(lr){2-3}
& Viewpoint 
& Framing 
& Overall   \\
& Error$\downarrow$
& Error$\downarrow$
& Error $\downarrow$ \\

\midrule
\multicolumn{4}{c}{\textit{World Model}} \\
\midrule
Veo3.1~\cite{veo} 
&1.351 & 0.749 &1.050\\
ViduQ2-Turbo~\cite{vidu} 
& 1.022 & 0.771  & 0.897 \\
Kling-V2.5~\cite{kling}     
& 1.051 & 0.733 &0.892\\
ReCamMaster~\cite{recammaster}     
& 0.755 & 0.720 &0.738\\
LingBot-World~\cite{lingbot-world}     
& 0.696 & 0.701 &{0.699}\\
\midrule
\multicolumn{4}{c}{\textit{Our Image Spatial Editing Model}} \\
\midrule
\rowcolor{gray!10} 
\textbf{SpatialEdit}           
& \textbf{0.243} & \textbf{0.527} & \textbf{0.385} \\

\bottomrule
\end{tabular}
\vspace{4pt}
\caption{\textbf{Performance comparison on proposed SpatialEdit-Bench.}}
\label{tab:world_model_space_edit}
\vspace{-25pt}
\end{table}
\section{More World Model Results}
We compare the performance of multiple video world models on precise camera viewpoint editing, including three closed-source models (ViduQ2-Turbo~\cite{vidu}, Kling-V2.5~\cite{kling}, Veo3.1~\cite{veo}) and two open-source models (LingBot-World~\cite{lingbot-world} and ReCamMaster~\cite{recammaster}), as summarized in~\cref{tab:world_model_space_edit}.

\section{Metrics in SpatialEdit-Bench}
\cref{alg:object_level} and~\cref{alg:camera_level} provide a clearer illustration of the metric calculation process used in SpatialEdit-Bench.
\begin{algorithm}
\caption{Object-Level Spatial Editing Evaluation}
\label{alg:object_level}
\begin{algorithmic}[2]

\renewcommand{\algorithmicrequire}{\textbf{Input:}}
\REQUIRE Images $\mathcal{I}_{\text{src}}, \mathcal{I}_{\text{pred}}$; BBox $b_{\text{gt}}$; Detector $\mathcal{Y}$; VLM $\mathcal{V}$

\renewcommand{\algorithmicrequire}{\textbf{Detection:}}
\REQUIRE $b_{\text{pred}} \gets \text{Detect}(\mathcal{Y}, \mathcal{I}_{\text{pred}})$

\STATE \textbf{Moving Score}:
\STATE \quad $\text{IoU} \gets \text{IoU}(b_{\text{gt}}, b_{\text{pred}})$
\STATE \quad $S_{\mathrm{oc}} \gets \mathcal{V}(\mathcal{I}_{\text{src}}, \mathcal{I}_{\text{pred}})$ \quad (object consistency)
\STATE \quad $\mathrm{MS} \gets \sqrt{\text{IoU} \cdot S_{\mathrm{oc}}}$

\STATE \textbf{Rotation Score}:
\STATE \quad $S_{\mathrm{view}} \gets \mathcal{V}(\mathcal{I}_{\text{pred}}, \theta, \phi)$ \quad (view correctness)
\STATE \quad $S_{\mathrm{cons}} \gets \mathcal{V}(\mathcal{I}_{\text{scr}}, \mathcal{I}_{\text{pred}})$ \quad (appearance consistency)
\STATE \quad $\mathrm{RS} \gets \sqrt{S_{\mathrm{view}} \cdot S_{\mathrm{cons}}}$

\STATE \quad Object Overall Score $\gets \frac{\text{MS} + \text{RS} }{2} $

\renewcommand{\algorithmicensure}{\textbf{Output:}}
\ENSURE $\mathrm{MS}$ (moving score), $\mathrm{RS}$ (rotation score), Object Overall Score

\end{algorithmic}
\end{algorithm}
\begin{algorithm}
\caption{Camera-Level Spatial Editing Evaluation}
\label{alg:camera_level}
\begin{algorithmic}[2]

\renewcommand{\algorithmicrequire}{\textbf{Input:}}
\REQUIRE Images $\mathcal{I}_{\text{src}}, \mathcal{I}_{\text{gt}}, \mathcal{I}_{\text{pred}}$; command $\Delta d$; detector $\mathcal{Y}$; VGGT model $f_{\text{VGGT}}$

\renewcommand{\algorithmicrequire}{\textbf{Pose Estimation:}}
\REQUIRE $(\widetilde{\mathcal{R}}_*, t_*) \gets f_{\text{VGGT}}(\mathcal{I}_{\text{src}}, \mathcal{I}_{\text{gt}}, \mathcal{I}_{\text{pred}})$ for $* \in \{\text{src},\text{gt},\text{pred}\}$

\renewcommand{\algorithmicrequire}{\textbf{Detection:}}
\REQUIRE $\mathcal{O}_* \gets \mathcal{Y}(\mathcal{I}_*)$ for $* \in \{\text{src},\text{gt},\text{pred}\}$

\STATE \textbf{Viewpoint Error:}
\STATE {Camera Center:} $\mathbf{C}_* \gets -\widetilde{\mathcal{R}}_*^{\top} t_*$
\STATE {Translation Error:} $\epsilon_{\text{xyz}} \gets \frac{\|\mathbf{C}_{\text{pred}}-\mathbf{C}_{\text{gt}}\|_2}{\|\mathbf{C}_{\text{gt}}-\mathbf{C}_{\text{src}}\|_2+\varepsilon}$
\STATE {Rotation Error:} $\epsilon_{\text{rot}} \gets \frac{1}{90} d_{\text{geo}}(\widetilde{\mathcal{R}}_{\text{pred}},\widetilde{\mathcal{R}}_{\text{gt}})$
\STATE $\mathrm{VE} \gets \tfrac{1}{2}(\epsilon_{\text{xyz}}+\epsilon_{\text{rot}})$

\STATE \textbf{Framing Error:}
\STATE {Ray:} $\mathbf{r}(u,v) \gets \text{norm}([(u-c_x)/f,(v-c_y)/f,1])$
\STATE {Match:} Hungarian matching between $\mathcal{O}_{\text{gt}}$ and $\mathcal{O}_{\text{pred}}$
\STATE \quad cost $c_{ij}=\angle(\mathbf{r}_i,\mathbf{r}_j)+\lambda|\ln(a_i/a_j)|$

\STATE $\epsilon_{\text{rag}} \gets \frac{1}{|\mathcal{M}|}\sum_{(i,j)\in\mathcal{M}}\arccos(\mathbf{r}_i^{\text{gt}}\!\cdot\!\mathbf{r}_j^{\text{pred}})\frac{180}{\pi}$

\STATE $\mathcal{M}_{\text{zoom}} \gets \text{Match}(\mathcal{O}_{\text{src}},\mathcal{O}_{\text{pred}})$
\STATE $\epsilon_{\text{zde}} \gets \mathbb{I}\!\left[\text{med}\!\left(\tfrac{1}{2}\ln\frac{|b_j^{\text{pred}}|}{|b_i^{\text{src}}|}\right)\!\cdot\!\Delta d>0\right]$

\STATE $\mathrm{FE} \gets \tfrac{1}{2}(\epsilon_{\text{rag}}+\epsilon_{\text{zde}})$

\STATE \quad Camera Overall Error $\gets \frac{\text{VE} + \text{FE}}{2}$

\renewcommand{\algorithmicensure}{\textbf{Output:}}
\ENSURE $\mathrm{VE}$ (viewpoint error), $\mathrm{FE}$ (framing error), Camera Overall Error

\end{algorithmic}
\end{algorithm}

\section{More Qualitative }
We provide more additional visualizations, as shown in~\cref{alg:object_level} and~\cref{alg:camera_level}.
For object-level manipulation tasks, we compare various open-source and closed-source editing models, while for camera-level manipulation tasks, we focus on evaluating the performance of different world models.
\begin{figure}[h]
  \centering
  \includegraphics[width=\linewidth]{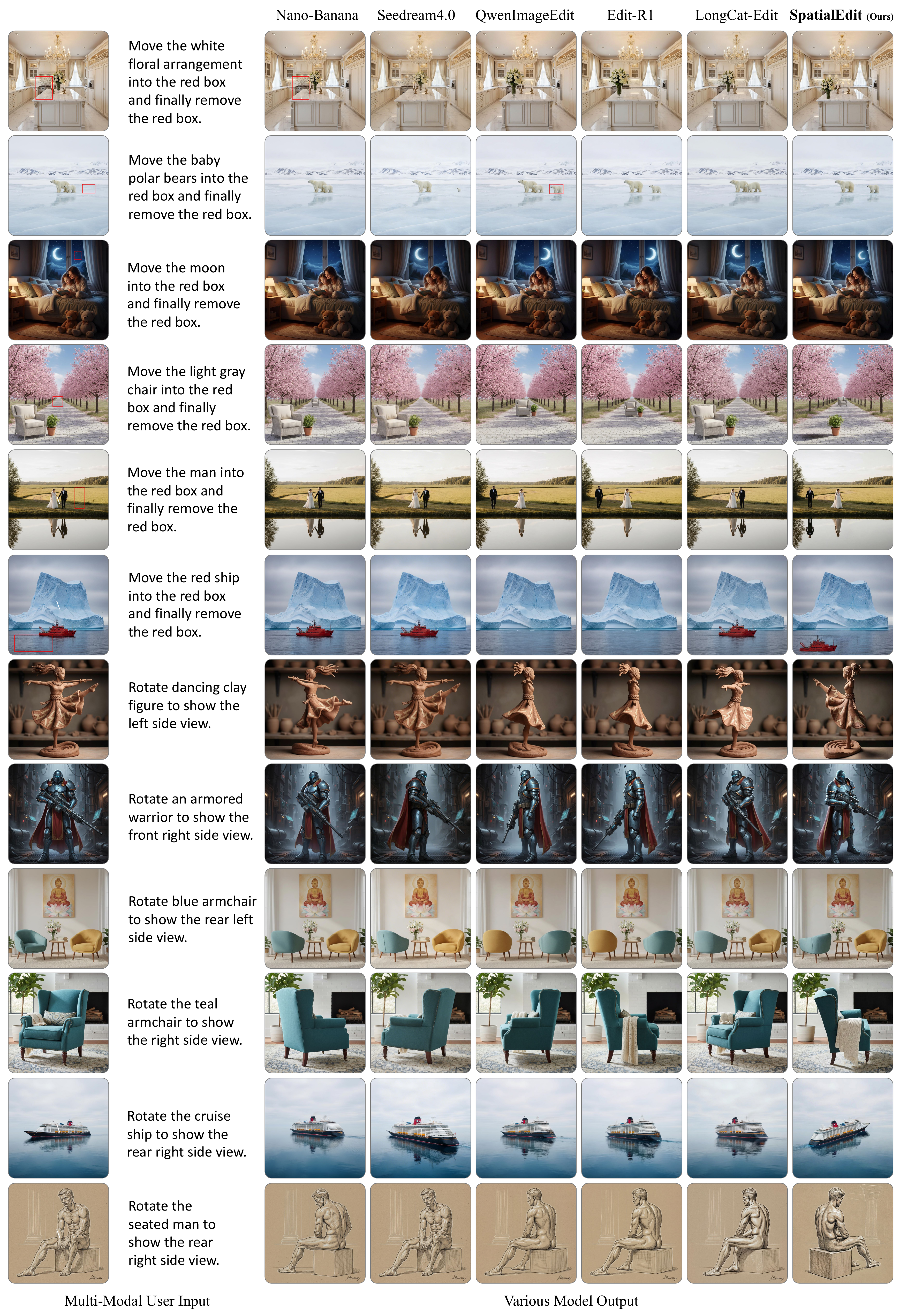}
  \caption{\textbf{Comparison of object-level manipulation across various methods}.
  }
  \label{fig:case_object_sup}
\end{figure}
\begin{figure}[h]
  \centering
  \includegraphics[width=\linewidth]{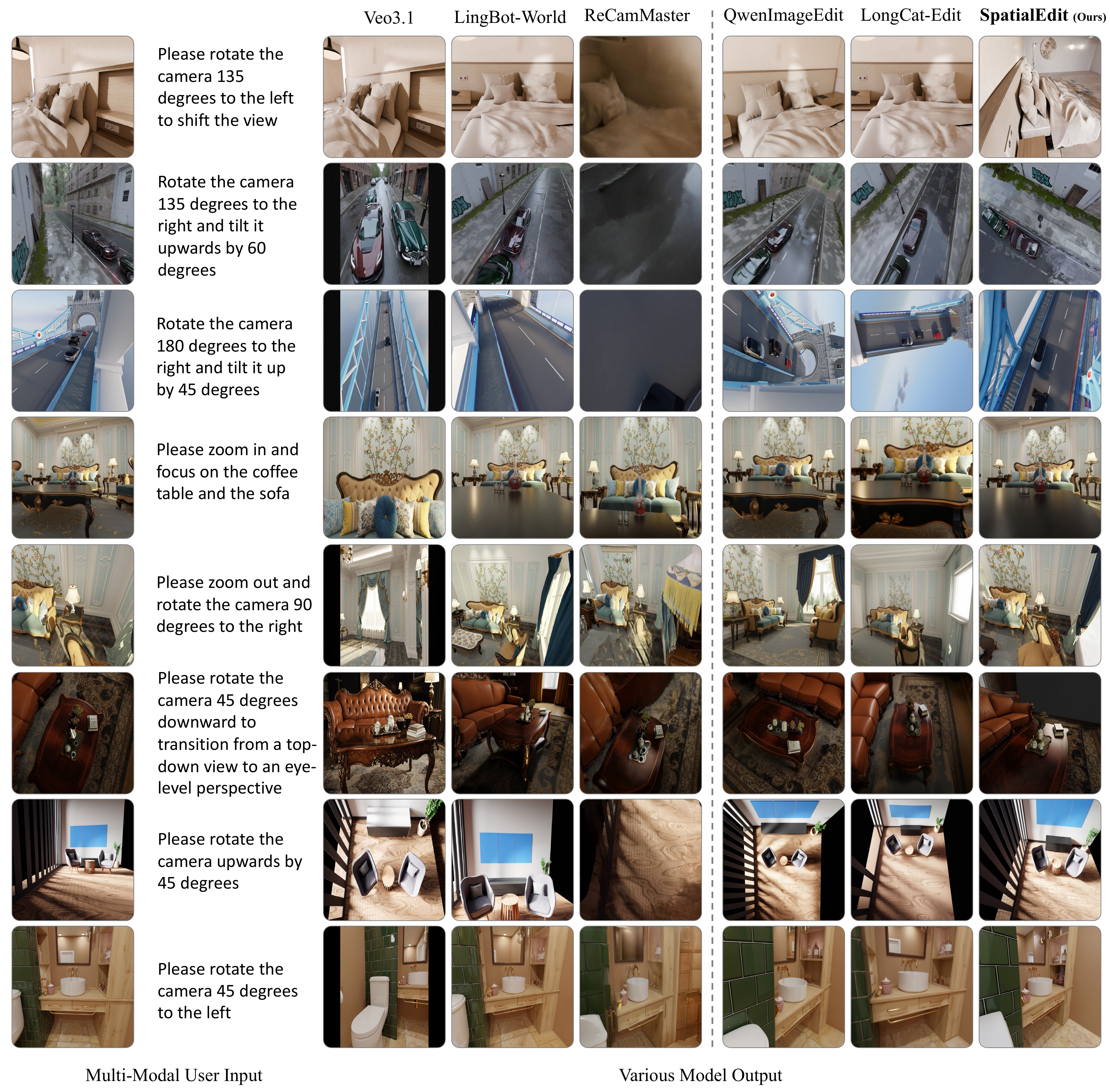}
  \caption{\textbf{Comparison of camera-level manipulation across various methods}.
  }
  \label{fig:case_camera_sup}
\end{figure}
\clearpage

\par\vfill\par


%
%
\bibliographystyle{splncs04}
\bibliography{main}
\end{document}